\documentclass{article}
\usepackage[utf8]{inputenc}
\usepackage{amsmath,amssymb,amsthm,graphicx,textcomp}
\usepackage{hyperref}
\theoremstyle{plain}
\newtheorem{theorem}{Theorem}
\newtheorem{proposition}[theorem]{Proposition}
\begin{document}
\title{Directional Non-Commutative Monoidal Structures with Interchange Law via Commutative Generators}
\author{Mahesh Godavarti}
\date{}
\maketitle

\begin{abstract}
We introduce a novel framework consisting of a class of algebraic structures that generalize one-dimensional monoidal systems into higher dimensions by defining per-axis composition operators subject to non-commutativity and a global interchange law. These structures, defined recursively from a base case of vector-matrix pairs, model directional composition in multiple dimensions while preserving structural coherence through commutative linear operators. 

We show that the framework that unifies several well-known linear transforms in signal processing and data analysis. In this framework, data indices are embedded into a composite structure that decomposes into simpler components. We show that classic transforms such as the Discrete Fourier Transform (DFT), the Walsh transform, and the Hadamard transform are special cases of our algebraic structure. The framework provides a systematic way to derive these transforms by appropriately choosing vector and matrix pairs. By subsuming classical transforms within a common structure, the framework also enables the development of learnable transformations tailored to specific data modalities and tasks.

\end{abstract}

\section{Introduction}
Classical monoidal structures, such as semigroups and matrix algebras, typically rely on a single associative binary operation. While sufficient for modeling one-dimensional sequences or temporal chains, they are inadequate for expressing multi-dimensional composition found in hierarchical data, spatial embeddings, or grid-structured information. In this work, we propose a new framework that allows for directional, non-commutative composition across arbitrary dimensions while satisfying the interchange law—a crucial structural constraint for preserving coherence between independent axes. This framework can be viewed simultaneously as an algebraic system and a representational model for structured data. 

\section{Background and Related Work}
\subsection{Monoidal Structures}
A monoidal structure consists of a set $S$ equipped with an associative binary operation $\ast: S \times S \to S$, and an identity element $e \in S$ such that $e \ast a = a \ast e = a$ for all $a \in S$. Examples include:
\begin{itemize}
\item $(\mathbb{N}, +, 0)$: natural numbers under addition.
\item $(\mathbb{R}^{n\times n}, \cdot, I)$: square matrices under multiplication.
\end{itemize}
Monoidal categories generalize these notions to categorical objects and morphisms, where composition is defined up to isomorphism. These structures are foundational in modern algebra, logic, and theoretical computer science [1].

\subsection{Bimonoidal Structures and Semirings}
A bimonoidal structure extends monoids by introducing two associative binary operations $\otimes$ and $\oplus$, subject to the interchange law: $(a \otimes b) \oplus (c \otimes d) = (a \oplus c) \otimes (b \oplus d)$. This is distinct from a semiring, where $\otimes$ must distribute over $\oplus$. Bimonoidal structures do not require distributivity, making them suitable for modeling independently composable dimensions without imposing algebraic symmetry [2,3].

\subsection{Multi-Monoidal and Categorical Generalizations}
Category theory extends these ideas through $n$-fold monoidal categories and double categories, where composition occurs along multiple independent axes. These frameworks formalize higher-dimensional algebraic reasoning [8]. Tensor networks used in quantum physics and deep learning also embody multi-axis contraction and composition [3,10]. Our construction shares this perspective but introduces symbolic, axis-specific, and non-commutative operations with strict interchange laws.

\subsection{Structured Embeddings and Symbolic Composition}
Recent works in machine learning have proposed directional and relative position embeddings that encode order sensitivity through group-theoretic and geometric transformations [5,6,9]. Our work generalizes this line of thought: instead of fixed or learned positional offsets, we encode transformation history through structured operator composition, allowing for deep recursive semantics and symbolic interaction.

\section{The One-Dimensional Case}
Let elements be pairs $(a, A)$, where:
\begin{itemize}
\item $a$ is a column vector (the vector component of $(a, A)$),
\item $A$ is an invertible matrix (the operator component).
\end{itemize}
We define the composition of two elements $(a, A)$ and $(b, B)$ as:
\[
(a, A) \circ (b, B) := \big(a + A\,b,\; A B\big)\,.
\]
This operation is associative; for any three elements $(a,A)$, $(b,B)$, $(c,C)$,
\[
((a, A) \circ (b, B)) \circ (c, C) = (a, A) \circ ((b, B) \circ (c, C))\,,
\]
and in general it is non-commutative:
\[
(a, A) \circ (b, B) \neq (b, B) \circ (a, A)\,.
\]
This non-commutativity stems from the directional nature of the transformation. In $(a,A)\circ(b,B)$, the operator $A$ from the first element acts on the vector $b$ of the second; reversing the order would have $B$ act on $a$ instead. Because $A B \neq B A$ in general, the result depends on the order of composition.

\section{Theorems and Proofs}
\subsection{Two-Dimensional Case}
We introduce two composition operations (one per axis) for a two-dimensional structure, subject to an interchange law:
\[
(a \circ_x b) \circ_y (c \circ_x d) = (a \circ_y c) \circ_x (b \circ_y d)\,. 
\]
We define elements in this 2D system as triples $(a, R_x^n, R_y^m)$, where:
\begin{itemize}
\item $a \in \mathbb{R}^d$ is a vector,
\item $R_x, R_y$ are linear operators (matrices) in $\mathbb{R}^{d\times d}$,
\item $n, m \in \mathbb{Z}$ are integer exponents indicating position along each axis.
\end{itemize}
\noindent \textbf{Composition rules:}
\begin{itemize}
\item \textbf{Horizontal composition ($\circ_x$):} $(a, R_x^n, R_y^m)\ \circ_x\ (b, R_x^k, R_y^m) = \big(a + R_x^n b,\; R_x^{\,n+k},\; R_y^m\big)\,. $
\item \textbf{Vertical composition ($\circ_y$):} $(a, R_x^n, R_y^m)\ \circ_y\ (b, R_x^n, R_y^q) = \big(a + R_y^m b,\; R_x^n,\; R_y^{\,m+q}\big)\,. $
\end{itemize}
To ensure the interchange law holds, we assume $R_x$ and $R_y$ commute. This guarantees that $R_x^i R_y^j = R_y^j R_x^i$ for all integers $i,j$.

\begin{theorem}[Interchange Law in 2D]\label{thm:2d}
If $R_x$ and $R_y$ commute (i.e., $R_x R_y = R_y R_x$), then the horizontal and vertical composition operations satisfy the interchange law. In particular, for elements 
\[ 
a=(u, R_x^n, R_y^m), \quad b=(v, R_x^k, R_y^m), \quad 
c=(w, R_x^n, R_y^q), \quad d=(z, R_x^k, R_y^q)\,, 
\] 
we have 
\[
(a \circ_x b)\ \circ_y\ (c \circ_x d) \;=\; (a \circ_y c)\ \circ_x\ (b \circ_y d)\,. 
\] 
\end{theorem}

\begin{proof}
By the composition rules, we have $a \circ_x b = \big(u + R_x^n v,\; R_x^{\,n+k},\; R_y^m\big)$ and $c \circ_x d = \big(w + R_x^n z,\; R_x^{\,n+k},\; R_y^q\big)$. Composing these results along the $y$-axis (which requires the $R_x$ exponents to match, as they do), we obtain 
\[
(a \circ_x b) \circ_y (c \circ_x d) \;=\; \Big(u + R_x^n v + R_y^m\big(w + R_x^n z\big),\; R_x^{\,n+k},\; R_y^{\,m+q}\Big)\,. 
\]
On the other hand, first composing vertically gives 
\[
a \circ_y c = \big(u + R_y^m w,\; R_x^n,\; R_y^{\,m+q}\big), \qquad 
b \circ_y d = \big(v + R_y^m z,\; R_x^k,\; R_y^{\,m+q}\big)\,. 
\] 
Now composing these along the $x$-axis yields 
\[
(a \circ_y c) \circ_x (b \circ_y d) = \Big((u + R_y^m w) + R_x^n\big(v + R_y^m z\big),\; R_x^{\,n+k},\; R_y^{\,m+q}\Big)\,. 
\]
Since $R_x R_y = R_y R_x$, the two results coincide. Thus, the combined composition is independent of whether horizontal or vertical composition is performed first, proving the interchange law.
\end{proof}

\subsection{General $D$-Dimensional Case}
Generalizing to $D$ dimensions, an element is represented as $(a,\; R_1^{n_1},\; R_2^{n_2},\; \dots,\; R_D^{n_D})$ with one operator $R_i$ per axis. The composition $\circ_i$ along axis $i$ is defined when all other indices $j \neq i$ are equal:
\[
(a, \,\dots,\; R_i^{\,n_i},\; \dots)\ \circ_i\ (b, \,\dots,\; R_i^{\,k_i},\; \dots) := \big(a + R_i^{\,n_i} b,\; \dots,\; R_i^{\,n_i+k_i},\; \dots\big)\,.
\]
If each pair of operators $R_i$ and $R_j$ commute, then interchange laws hold between every pair of dimensions. In other words, the final result of composing a collection of elements is independent of the order in which multi-axis compositions are applied.

Note that for $D=1$ (a single axis), the interchange law is trivial, so each element may have its own operator (e.g., an element $(v_i, R_i)$ can have a distinct $R_i$). For $D>1$, by contrast, we impose a fixed operator $R_i$ for each dimension $i$ across all elements, to ensure a consistent interchange law holds across dimensions.

\begin{proposition}\label{prop:Ddim}
For any $D \ge 2$, if $R_1, R_2, \ldots, R_D$ commute pairwise, then the $D$ composition operations $\circ_1,\circ_2,\ldots,\circ_D$ satisfy all pairwise interchange laws. Equivalently, any two ways of iteratively composing $D$-dimensional elements (differing only by the order of axis composition) yield the same final element.
\end{proposition}

\begin{proof}
The case $D=2$ is exactly Theorem~\ref{thm:2d}. For $D>2$, we proceed by induction. Assume the claim holds for all structures up to $D-1$ dimensions. Consider a collection of elements in $D$ dimensions. We can group the elements into $(D-1)$-dimensional slices indexed by the $D$-th coordinate. By the inductive hypothesis, the composition within each slice is well-defined (order-independent). Moreover, since $R_D$ commutes with every other $R_i$ by assumption, composing along the $D$-th axis can be interleaved with compositions along the other axes without affecting the outcome. Hence, all interchange relations involving the $D$-th axis also hold, establishing the property for $D$ dimensions.
\end{proof}

In addition to the interchange law, the above structures satisfy several intuitive algebraic properties. Each composition operator $\circ_i$ is associative (since it is defined via vector addition and operator multiplication, both of which are associative). Different directional compositions are generally non-commutative with each other, reflecting the order-sensitivity of multi-axis composition when interchange is not enforced. The construction is closed under growing the operator exponents (applying successive transformations accumulates the exponent values along each axis). Finally, by construction and the commutativity assumptions on the $R_i$, the interchange law holds for every pair of axes, as proven above.

\section{Composition of Embeddings}
In one dimension, each element can be represented as a tuple $e_i = (v_i, R_i)$. For a sequence $x_1, x_2, \ldots, x_T$, we compose the element embeddings sequentially to encode the entire sequence:
\[
E = e_1 \circ (e_2 \circ (e_3 \dots (e_{T-1} \circ e_T)))\,.
\]
Expanding the expression, we obtain
\[
E = \Big(v_1 + R_1 v_2 + R_1 R_2 v_3 + \cdots + R_1 R_2 \cdots R_{T-2} R_{T-1} v_T,\; R_1 R_2 \cdots R_{T-2} R_{T-1}\Big)\,.
\]
 
In two dimensions, an element can be written as $e_{ij} = (v_{ij}, R_x, R_y)$ for positions $1 \le i \le H$ and $1 \le j \le W$. Composing all elements of an $H \times W$ grid yields:
\[
E = \bigcirc_{i=1}^H \bigcirc_{j=1}^W e_{ij} = \Big(\sum_{i=1}^H \sum_{j=1}^W R_x^{\,i-1} R_y^{\,j-1} v_{ij},\; R_x^H R_y^W\Big)\,.
\]
Here we use two composition operators, $\circ_x$ and $\circ_y$, corresponding to moves in the horizontal ($x$) and vertical ($y$) directions, respectively. 

This approach extends naturally to higher dimensions by introducing one composition operator per additional axis. In such cases, the interchange law generalizes to all pairs of axes, provided that the associated operators commute pairwise.

\section{Discrete Fourier Transform as a Special Case}
In this section, we demonstrate how the discrete Fourier transform (DFT) can be obtained as a special case of the embedding composition framework. We first consider the one-dimensional (1D) DFT and then generalize the construction to two dimensions.

\subsection{One-Dimensional Case}
Consider a sequence of $n$ input embeddings $(v_i, R)$ for $i=1,2,\ldots,n$, all sharing the same linear transformation $R$. It is a property of sequential embeddings that their composition yields a single embedding $(V, R^n)$, where the combined vector $V$ is given by the sum of the intermediate vectors transformed appropriately:
\begin{equation}\label{eq:embed-sum}
    V \;=\; \sum_{i=1}^{n} R^{\,i-1}\, v_i~.
\end{equation}
We will choose $R$ and $v_i$ such that $V$ encodes the DFT of a given length-$n$ signal $a_1, a_2, \ldots, a_n$. 

Let $R$ be a real $2n \times 2n$ block-diagonal matrix consisting of $n$ rotation blocks of size $2\times2$. Indexing the blocks by $k=0,1,\ldots,n-1$, the $k$-th diagonal block $R_k$ is defined as 
\[
    R_k \;=\; \begin{pmatrix}
        \cos\!\frac{2\pi k}{n} & -\,\sin\!\frac{2\pi k}{n} \\[6pt]
        \sin\!\frac{2\pi k}{n} & \cos\!\frac{2\pi k}{n} 
    \end{pmatrix},
\] 
which corresponds to multiplication by the complex number $e^{\,j 2\pi k/n}$ (as a rotation in the real plane). Thus 
\[ 
    R \;=\; \mathrm{diag}(R_0,\,R_1,\,\dots,\,R_{n-1})~,
\] 
and note that $R^n = I_{2n}$ (each $R_k^n$ is a rotation by $2\pi k$, yielding the identity). For the input vectors, we take 
\[
    v_i \;=\; \begin{bmatrix} a_i \\[3pt] 0 \\[3pt] a_i \\[3pt] 0 \\ \vdots \\[3pt] a_i \\[3pt] 0 \end{bmatrix} \in \mathbb{R}^{2n},
\] 
i.e. $v_i$ is formed by repeating the two-dimensional column vector $(a_i,\;0)^T$ across each of the $n$ blocks. Substituting these choices into \eqref{eq:embed-sum}, we obtain
\begin{equation}\label{eq:V-components}
    V \;=\; \sum_{i=1}^{n} R^{\,i-1} v_i~.
\end{equation}
Because $R$ is block-diagonal, the effect of $R^{\,i-1}$ on $v_i$ is to rotate each $2\times2$ block of $v_i$ by an angle of $(i-1)\frac{2\pi k}{n}$ in the $k$-th block. This yields, for each block $k$, a two-component contribution 
\[
    R^{\,i-1} v_i \Big|_{\text{block }k} \;=\; 
    \begin{pmatrix}
      a_i \cos\!\dfrac{2\pi k (i-1)}{n} \\[8pt]
      a_i \sin\!\dfrac{2\pi k (i-1)}{n}
    \end{pmatrix}\!. 
\] 
Summing over $i=1$ to $n$ as in \eqref{eq:V-components}, the $k$-th block of the resulting vector $V$ is 
\begin{equation}\label{eq:Vk}
    V\Big|_{\text{block }k} \;=\; 
    \begin{pmatrix}
      \sum_{i=1}^n a_i \cos\!\dfrac{2\pi k (i-1)}{n} \\[2.2em]
      \sum_{i=1}^n a_i \sin\!\dfrac{2\pi k (i-1)}{n}
    \end{pmatrix}, \qquad k = 0,1,\ldots,n-1~.
\end{equation}
Interpreting each pair as the real and imaginary parts of a complex number, we recognize \eqref{eq:Vk} as the components of the length-$n$ DFT of the sequence $\{a_i\}$. In particular, if we define 
\[ 
    X_k \;=\; \sum_{i=1}^n a_i\, e^{\,j \frac{2\pi}{n}(i-1)k}~, \qquad k=0,\ldots,n-1,
\] 
then $V\big|_{\text{block }k} = \begin{pmatrix} \Re(X_k) \\[3pt] \Im(X_k) \end{pmatrix}$. Thus, the vector $V$ produced by the embedding construction exactly encodes all $n$ DFT coefficients $X_k$. Moreover, since $R^n$ is the identity, the overall transformation of the embedding after $n$ steps is trivial, and the output $(V, R^n)$ effectively yields just the vector $V$ containing the DFT results.

\subsection{Two-Dimensional Case}
The above approach extends naturally to the two-dimensional (2D) DFT. While the discussion here focuses on the 2D DFT, the extension to the $N$-dimensional case is straightforward by applying the 1D DFT along each axis independently. 

Consider an $n\times n$ data array $a_{i,j}$ (with $i,j=1,\ldots,n$). We can recover its 2D DFT by applying the one-dimensional embedding method along each dimension in succession, using the same rotation matrix $R$ for both the row and column directions.

First, we perform the 1D embedding-based transform along each row $i$. For each fixed $i$, we treat $(a_{i,1}, a_{i,2}, \dots, a_{i,n})$ as an input sequence and apply the above construction (with the same $R$) across the index $j$. Because $R^n = I$, this yields an output vector $W_i$ for each row $i$, whose $k$-th block is 
\[
    W_i\Big|_{\text{block }k} \;=\; 
    \begin{pmatrix}
      \sum_{j=1}^n a_{i,j} \cos\!\dfrac{2\pi k (j-1)}{n} \\[1.2em]
      \sum_{j=1}^n a_{i,j} \sin\!\dfrac{2\pi k (j-1)}{n}
    \end{pmatrix}
    \;=\;
    \begin{pmatrix} \Re\!\big(X_{i,k}^{\text{(row)}}\big) \\[3pt] \Im\!\big(X_{i,k}^{\text{(row)}}\big) \end{pmatrix}\!,
\] 
where $X_{i,k}^{\text{(row)}} \;=\; \sum_{m=1}^n a_{i,m}\, \exp\!\Big(j \frac{2\pi}{n}(m-1)k\Big)$ is the 1D DFT of the $i$-th row. In other words, after this step each row $i$ has been transformed independently into its Fourier spectrum $\{X_{i,k}^{\text{(row)}}: k=0,\ldots,n-1\}$.

In the second step, we treat each column of the intermediate results $W_i$ as an $n$-length sequence in the vertical index $i$, and apply the same embedding construction along that axis. Specifically, for each fixed frequency index $k$ (one of the $n$ columns of the intermediate spectra), we apply the embedding method with $R$ across $i=1,\ldots,n$. By exactly the same reasoning as in the 1D case, the outcome for each column $k$ is a vector $Y^{(k)}$ whose $p$-th block ($p=0,\ldots,n-1$) corresponds to the complex number 
\[
    Y_{p,k} \;=\; \sum_{i=1}^n X_{i,k}^{\text{(row)}} \, e^{\,j \frac{2\pi}{n}(i-1)p}~,
\] 
which is precisely the 1D DFT (along the $i$-direction) of the sequence $\{X_{i,k}^{\text{(row)}}\}_{i=1}^n$. Substituting the expression for $X_{i,k}^{\text{(row)}}$ from above, we obtain 
\begin{equation}\label{eq:2ddft}
    Y_{p,k} \;=\; \sum_{i=1}^n \sum_{m=1}^n a_{i,m}\, \exp\!\Big(j \frac{2\pi}{n}\big[(m-1)k + (i-1)p\big]\Big)~,
\end{equation}
for each $p, k = 0,1,\ldots,n-1$. Here $Y_{p,k}$ is exactly the $(p,k)$ entry of the 2D DFT of the array $a_{i,j}$. We recognize the double sum in \eqref{eq:2ddft} as separable, i.e. 
\[ 
    Y_{p,k} = \sum_{i=1}^n \Big(\sum_{m=1}^n a_{i,m} e^{\,j \frac{2\pi}{n}(m-1)k}\Big) e^{\,j \frac{2\pi}{n}(i-1)p}~, 
\] 
showing explicitly that the 2D transform can be carried out by first performing $n$ independent 1D DFTs along the rows (inner sum) and then $n$ independent 1D DFTs along the columns (outer sum). This two-step procedure is exactly what the sequential embedding with the same rotation $R$ along both axes achieves. In summary, by composing the 1D embedding method along the horizontal and vertical directions, we recover the full 2D DFT of $a_{i,j}$.

\section{Hadamard Transform as a Special Case}

We now give a full construction showing that the Hadamard transform \cite{ref13} can be written as:
\[
H_n x = \sum_{i=1}^n R^{i-1} v_i
\]
for a specific choice of matrix \( R \in \mathbb{R}^{n \times n} \) and vectors \( v_i \in \mathbb{R}^n \). We assume \( n = 2^m \).

\paragraph{Definition of \( R \):}
Let \( R = \mathrm{diag}(1, -1, 1, -1, \ldots, 1, -1) \), an \( n \times n \) diagonal matrix alternating between \( 1 \) and \( -1 \). Then \( R^2 = I \).

\paragraph{Definition of \( v_i \):}
Let \( x = (x_1, x_2, \ldots, x_n) \in \mathbb{R}^n \). Construct \( v_i \in \mathbb{R}^n \) as follows:
\[
v_i = \begin{cases}
x_i \cdot (e_i + e_{i + n/2}) & \text{if } 1 \leq i \leq n/2 \\
x_i \cdot (e_{i - n/2} - e_i) & \text{if } n/2 < i \leq n
\end{cases}
\]
where \( e_j \) is the \( j \)-th standard basis vector of \( \mathbb{R}^n \).

\paragraph{Base Case:}
When \( n = 2 \), we have:
\[
H_2 = \begin{pmatrix} 1 & 1 \\ 1 & -1 \end{pmatrix},\quad R = \begin{pmatrix} 1 & 0 \\ 0 & -1 \end{pmatrix}
\]
\[
v_1 = x_1 \begin{pmatrix} 1 \\ 1 \end{pmatrix},\quad v_2 = x_2 \begin{pmatrix} 1 \\ -1 \end{pmatrix}
\Rightarrow R^0 v_1 + R^1 v_2 = v_1 + R v_2 = \begin{pmatrix} x_1 + x_2 \\ x_1 - x_2 \end{pmatrix} = H_2 x
\]

\paragraph{Inductive Step:}
Assume the claim holds for \( n = 2^k \), and consider \( n' = 2n = 2^{k+1} \). Write:
\[
x = \begin{pmatrix} x^{(1)} \\ x^{(2)} \end{pmatrix},\quad x^{(1)}, x^{(2)} \in \mathbb{R}^n
\]
Then by Sylvester's construction:
\[
H_{2n} x = \begin{pmatrix} H_n & H_n \\ H_n & -H_n \end{pmatrix} \begin{pmatrix} x^{(1)} \\ x^{(2)} \end{pmatrix}
= \begin{pmatrix} H_n(x^{(1)} + x^{(2)}) \\ H_n(x^{(1)} - x^{(2)}) \end{pmatrix}
\]
Using the inductive assumption, both terms can be written as sums of \( R_n^{i-1} v_i \). Thus, the inductive step holds.

The extension to 
$N$-dimensional Hadamard transforms follows naturally via tensor products of the 1D transform along each axis.

\section{Walsh Transform as a Special Case}
The \emph{Walsh} transform \cite{ref14} is closely related to the Hadamard transform.  A Walsh matrix $W_n$ of order $n=2^k$ is obtained by reordering the rows (or columns) of the Sylvester Hadamard matrix so that the number of sign changes (sequency) increases row by row.  Equivalently, one can define $W_n = P H_n$ where $P$ is a fixed permutation matrix that rearranges indices according to the binary Gray code or sequency order.  We will show how Walsh fits into our framework by incorporating this permutation.

Formally, let $H_n$ be the standard Hadamard matrix (Sylvester order) and let $P$ be the Gray-code permutation matrix such that $W_n = P H_n$.  Equivalently, one can also write $W_n x = H_n(Px)$ for any input $x$.  Since we have already embedded $H_n$, we can embed $W_n$ as follows.  Given input $x$, first apply the permutation: consider $x' = P x$.  Then $H_n x'$ can be written as $\sum_i R^{\,i-1}v'_i$ by our previous construction (with $R,v_i$ for $H_n$).  Finally apply the same permutation to the output.  In other words, $W_n x = P\,H_n\,(P^{-1}x)$. 

Concretely, if $\{R,v_i\}$ represent $H_n$, then one can define $R' = P\,R\,P^{-1}$ (a conjugated block-diagonal operator) and $v'_i = P\,v_i$ (permuted embeddings).  Then 
\[
\sum_i {R'}^{\,i-1}v'_i 
= P \Bigl(\sum_i R^{\,i-1}v_i\Bigr) 
= P\,H_n(P^{-1}x) 
= W_n x.
\]
Thus Walsh transforms are realized by introducing a permutation operator in the framework.  In practice, one often simply notes that the Walsh matrix can be obtained by reordering $H_n$ in Gray-code (sequency) order, and that this permutation is itself a linear operator.  Therefore the compositional embedding for Walsh is obtained by combining the previously defined embedding for $H_n$ with a permutation. 

Similarly, the $N$-dimensional Walsh transform can be constructed by applying the 1D Walsh transform along each coordinate axis, making the extension immediate within this framework.

\section{Conclusion}
We have presented a framework for multi-dimensional, non-commutative composition that preserves structural consistency through commuting transformations. By introducing distinct composition operators for independent axes and enforcing an interchange law via commutative linear operators, we generalized classical monoids to a higher-dimensional setting. The resulting algebraic structure maintains associativity along each axis and yields coherent global compositions that are independent of axis ordering. Our construction provides a new perspective on positional embeddings in machine learning and symbolic sequences, as demonstrated by deriving the discrete Fourier transform as a special case of the embedding composition method. 

We have shown that the DFT, the Hadamard transform, and the Walsh transform each admit a representation within the compositional embedding framework.  For the DFT, we constructed a $2N\times 2N$ block-diagonal rotation matrix $R$ and embedding vectors $\{v_i\}$ which reproduce the real and imaginary components of the DFT outputs.  For the Hadamard transform, we used Sylvester’s recursive definition to inductively build $R$ and $v_i$ so that $H_nx = \sum_iR^{i-1}v_i$ for all $n=2^k$.  Finally, the Walsh transform (sequency-ordered Hadamard) is obtained by applying a Gray-code permutation to the Hadamard case, which we incorporated by conjugating by a permutation matrix.  

From a broad mathematical standpoint, this framework initiates the development of a general theory of structured compositional representations—one that integrates algebraic constructions, geometric intuitions, and computational realizations. By revealing deep connections between discrete transforms and compositional algebraic operations, this approach opens new possibilities for unifying diverse transformation techniques under a common formalism. Such a theory has the potential to impact a wide range of disciplines, including signal processing, machine learning, and scientific computing, by providing a principled foundation for building learnable transformations, and interpretable and efficient computational models.

\section*{Acknowledgements}
I extend my sincere gratitude to Dr. S. Prem Kumar for invaluable discussions and insights that significantly contributed to the development of the ideas presented in this work.

\end{document}